\documentclass[11pt,a4paper]{article}
\usepackage[hyperref]{emnlp-ijcnlp-2019}
\usepackage{times}
\usepackage{latexsym}
\usepackage{graphicx}
\usepackage{url}
\usepackage{adjustbox}
\aclfinalcopy 
\usepackage{adjustbox}
\usepackage{multirow}
\usepackage{amsmath,amssymb}
\usepackage{arydshln}

\usepackage{dsfont}

\title{A Stack-Propagation Framework with Token-Level Intent Detection for Spoken Language Understanding }

\author{Libo Qin, Wanxiang Che\thanks{* Email corresponding.}, Yangming Li, Haoyang Wen, Ting Liu \\
	Research Center for Social Computing and Information Retrieval \\
	Harbin Institute of Technology, China \\
	{\tt \{lbqin,car,yangmingli,hywen,tliu\}@ir.hit.edu.cn}	
}

\date{}

\begin{document}
\maketitle
\begin{abstract}
Intent detection and slot filling are two main tasks for building a spoken language understanding (SLU) system. 
The two tasks are closely tied and the slots often highly depend on the
intent. 
In this paper, we propose a novel framework for SLU to better incorporate the intent information, which further guides the slot filling.
In our framework, we adopt a joint model with Stack-Propagation which can directly use the intent information as input for slot filling, thus to capture the intent semantic knowledge.
In addition, to further alleviate the error propagation, we perform the token-level intent detection for the Stack-Propagation framework.
Experiments on two publicly
datasets show that our model achieves the state-of-the-art 
performance and outperforms other previous methods by a large 
margin. 
Finally, we use the Bidirectional Encoder 
Representation from Transformer (BERT) model in our framework, which further boost our performance in SLU task.
\end{abstract}

\section{Introduction}
Spoken language understanding (SLU) is a critical component in task-oriented dialogue systems. It usually consists of intent detection
to identify users' intents and slot filling task to extract semantic constituents from the natural language utterances \cite{tur2011spoken}. 
As shown in Table~\ref{example1}, given a
movie-related utterance ``\textit{watch action movie}'', there are different slot labels for each token and an intent for the whole utterance.

Usually, intent detection and slot filling are implemented separately. But intuitively, 
these two tasks are not independent and the slots often highly depend on the
intent \cite{goo2018slot}. For example, if the intent of a utterance is
\texttt{WatchMovie}, it is more likely to contain the slot \texttt{movie\_name} rather
than the slot \texttt{music\_name}. 
Hence, it is promising to incorporate the intent information to guide the slot filling.
\begin{table}[t]
	\centering
	\begin{adjustbox}{width=0.48\textwidth}
		\tiny
		
		\begin{tabular}{|l|c|c|c|}
			\hline
			\textbf{Sentence} & \textit{watch} & \textit{action} & \textit{movie} \\
			\hline
			\textbf{Gold Slots} & \texttt{O} & \texttt{B-movie\_name} & \texttt{I-movie\_name}  \\
			\hline      
			\textbf{Gold Intent} & \multicolumn{3}{c|}{\texttt{WatchMovie}}\\
			\hline    
		\end{tabular}
		
	\end{adjustbox}
	
	\caption{An example  with intent and slot annotation (BIO format), which indicates the slot of movie name from an utterance with an intent \texttt{WatchMovie}.}
	\label{example1}
\end{table}

Considering this strong correlation between
the two tasks, some joint models are proposed based on the multi-task learning framework \cite{zhang2016joint,hakkani2016multi,liu2016attention} and all these models outperform the pipeline models via mutual enhancement between two tasks. However, their work just modeled the relationship between intent and slots by sharing parameters.
Recently, some work begins to model the intent information for slot filling explicitly in joint model.
\citet{goo2018slot} and \citet{li2018self} proposed the gate mechanism to explore incorporating the intent information for slot filling.
Though achieving the promising performance, their models still suffer from two issues including: (1) They all adopt the gate vector to incorporate the intent information. 
In the paper, we argue that it is risky to simply rely on the gate function
to summarize or memorize the intent information.
Besides,  the interpretability of how the intent information guides slot filling procedure is still weak due to the interaction with hidden vector between the two tasks. 
(2) The utterance-level intent information they use for slot filling may mislead the prediction for all slots in an utterance if the predicted utterance-level intent is incorrect.
\begin{figure}[t!]
	\centering
	\includegraphics[scale=0.33]{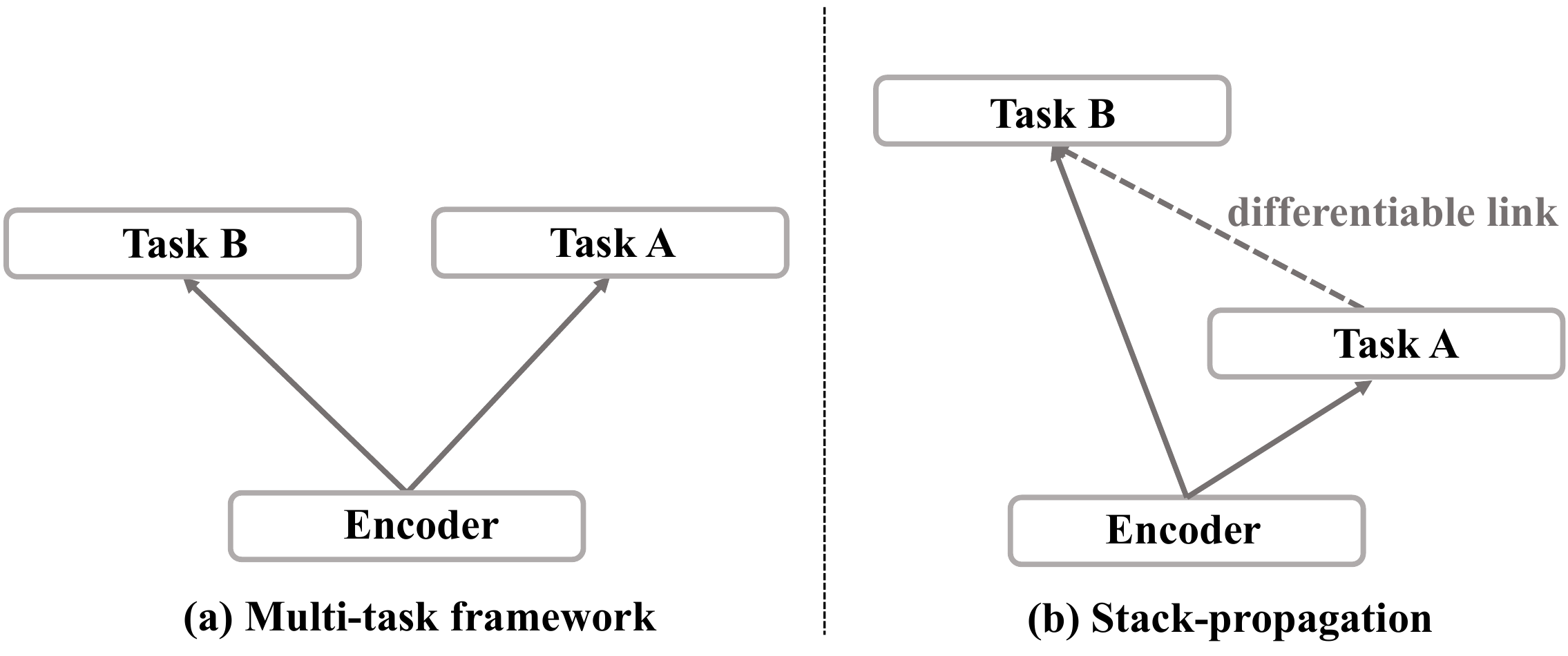}
	\caption{Multi-task framework vs. Stack-Propagation.}\label{fig:stack}
\end{figure}

In this paper, we propose a novel framework to address both two issues above.
For the first issue, inspired by the Stack-Propagation which was proposed by \newcite{P16-1147} to leverage the POS 
tagging features for parsing and achieved good performance,
we propose a joint model with Stack-Propagation for SLU tasks. Our framework directly use the output of the intent detection as the input for slot filling to better guide the slot prediction process. In addition, the framework make it easy to design oracle intent experiment to intuitively show how intent information enhances slot filling task.
For the second issue, we perform a token-level intent prediction in our framework, which can provide the token-level intent information for slot filling.
If some token-level intents in the utterance are predicted incorrectly, other correct token-level intents will still be useful for the corresponding slot prediction.
In practice,
we use a self-attentive encoder for intent detection to capture the contextual 
information at each token and hence predict an intent label at 
each token.
The intent of an utterance is computed by voting from predictions at each token of the utterance. This token-level prediction, like ensemble neural networks \cite{lee2016stochastic}, reduces the predicted variance to improve the performance of intent detection.
And it fits better in our Stack-Propagation framework, where intent detection
can provide token-level intent features and retain more useful intent information for slot filling.

We conduct experiments on two benchmarks SNIPS \cite{coucke2018snips} and ATIS \cite{goo2018slot} datasets. The results of both experiments show the effectiveness of our framework by outperforming the current state-of-the-art methods by a large margin. 
Finally, Bidirectional Encoder Representation from Transformer
\cite[BERT]{devlin2018bert}, 
as the pre-trained model, is used to further boost the performance of our model. 

To summarize, the contributions of this work are as follows:
\begin{itemize}
	\item We propose a Stack-Propagation framework in SLU task, which
	 can better incorporate the intent semantic knowledge to guide the slot filling and make our joint model more interpretable.

	\item We perform the token-level intent detection for Stack-Propagation framework, which improves the intent detection performance and further alleviate the error propagation. 
	
	\item We present extensive experiments demonstrating the benefit of our proposed framework. Our experiments on two publicly available datasets show substantial improvement and our framework achieve the state-of-the-art performance.
	
	\item  We explore and analyze the effect of incorporating BERT in SLU tasks.
\end{itemize}

For reproducibility, our code for this paper is publicly available at
\url{https://github.com/LeePleased/StackPropagation-SLU}.

\section{Background}
In this section, we will describe the formulation definition for intent detection
and slot filling, and then we give a brief
description of the multi-task framework and the joint model with Stack-Propagation framework.

\subsection{Intent Detection and Slot Filling}
Intent detection can be seen as a classification problem to decide the intent label $o^{I}$ 
of an utterance. 
Slot filling is a sequence labeling task that maps an input word sequence
$\mathbf{ x}$ = $({x}_{1} ,\dots,{x}_{T} )$ to slots sequence $\mathbf{o}^{S}$ = $({o}_{1}^{S} , \dots , {o}_{T}^{S})$.

\subsection{Multi-task Framework vs. Stack-Propagation}
For two correlative tasks \textit{Task A} and \textit{Task B}, multi-task framework which is shown in Figure~\ref{fig:stack}~(a) can learn the correlations between these two tasks by the shared encoder. 
However, the basic multi-task framework cannot provide features from up-stream task to down-stream task explicitly.
The joint model with
Stack-Propagation framework which is shown in Figure~\ref{fig:stack}~(b) can mitigate the shortcoming.
In this figure, \textit{Task B} can leverage features of \textit{Task A} without breaking the differentiability in
Stack-Propagation framework and promote each other by joint learning at the same time.
\begin{figure} [t]
	\centering
	\includegraphics[scale=0.135]{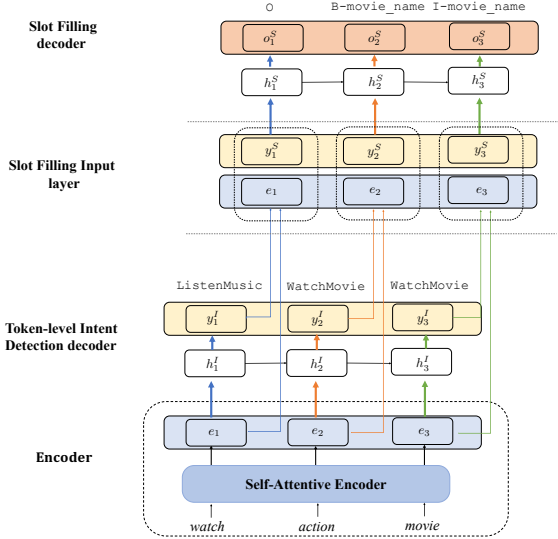}
	\caption{Illustration of our Stack-Propagation framework for joint intent detection and slot filling. It consists of one shared self-attentive encoder and two decoders. The output distribution of intent detection network and the representations from encoder are concatenated as the input for slot filling.}
	\label{fig:framework}
\end{figure}
\section{Approach}
In this section, we will describe our Stack-Propagation framework for SLU task.
The architecture of our framework is demonstrated in Figure~\ref{fig:framework}, 
which consists of an encoder and two decoders. 
First, the encoder module uses one shared self-attentive encoder to represent an utterance, which can grasp the shared knowledge between two tasks.
Then, the intent-detection decoder performs a token-level intent detection.
Finally, our Stack-Propagation framework leverages the explicit token-level intent information for slot filling 
by concatenating the output of intent detection decoder and the representations from encoder as the input for slot filling decoder. 
Both intent detection and slot filling are optimized simultaneously via a joint learning scheme.

\subsection{Self-Attentive Encoder}
In our Stack-Propagation framework, intent detection task and slot filling task share one encoder, 
In the self-attentive encoder, 
we use BiLSTM with self-attention mechanism to 
leverage both advantages of temporal features and contextual information, which 
are useful 
for sequence labeling tasks \cite{P18-1135,yin2018deep}.

The BiLSTM  \cite{hochreiter1997long} reads input utterance $\bf{X}$ = (${\bf{x}}_{1}, {\bf{x}}_{2},.., {\bf{x}}_{T}$)  ($T$ is the 
number of tokens in the input utterance)  forwardly and backwardly to produce context-sensitive  hidden states $\bf{H}$ =  $(\mathbf{h}_{1}, \mathbf{h}_2, ..., \mathbf{h}_{T})$
by repeatedly applying the recurrence 
$\mathbf{h}_{i}=\text{BiLSTM}\left( \phi ^{\text{emb}}\left( x_{i}\right) , \mathbf{h}_{i-1}\right)$.

Self-attention is a very effective 
method of leveraging context-aware features over 
variable-length sequences for natural language 
processing tasks \cite{tan2018deep,P18-1135}. 
In our case, we use  self-attention mechanism to capture 
the contextual information for each token.
In this paper, we adopt \citet{NIPS2017_7181}, where
we first map the matrix of input vectors $\bf{X}$ $\in$ $\mathbb{R}^{T\times d}$ 
($d$ represents the 
mapped dimension) to queries (${\mathbf{Q}}$), keys (${\mathbf K}$) and values 
(${\mathbf V}$) matrices by using different linear projections
and the self-attention output 
{\bf{C}} $\in$ $\mathbb{R}^{T\times d}$  is a weighted sum of values:
\begin{equation}
{\bf{C}} = \operatorname { softmax } \left( \frac { \mathbf Q \mathbf K^\top } { \sqrt { d _ { k } } } \right) \mathbf V.
\end{equation}

After obtaining the output of self-attention and 
BiLSTM. We concatenate these two representations as 
the final encoding representation:
\begin{equation}
\mathbf{ E} =\mathbf{ H } \oplus\mathbf{C},
\end{equation}
where {\bf{E}} $\in$ $\mathbb{R}^{T\times 2d}$ and $\oplus$ is concatenation operation.

\subsection{Token-Level Intent Detection Decoder}

In our framework, we perform a token-level intent detection, which can provide 
token-level intent features for slot filling, different from regarding 
the intent detection task as the sentence-level classification problem 
\cite{liu2016attention}.
The token-level intent detection method can be formalized as a sequence labeling problem that
maps a input word sequence $\mathbf{ x}$ = $(x_{1} ,...,x_{T} )$ to
sequence of intent label $\mathbf{o}^{I}$ = $(o_{1}^{I} , ... , o_{T}^{I})$. 
In training time, we set the sentence's intent label as every token's gold intent label.
The final intent of an utterance is computed by voting from predictions 
at each token of the utterance.

The self-attentive encoder generates a 
sequence of contextual representations $\mathbf{E}= (\mathbf{e}_{1},...,\mathbf{e}_T)$ and each 
token can grasp the whole contextual information by self-attention mechanism. We use a unidirectional LSTM as the intent detection network. 
At each decoding step $i$, the decoder state ${\bf{h}}_{i}^{I}$ is 
calculated by previous decoder state ${\bf{h}}_{i-1}^{I}$, the previous 
emitted intent label distribution ${\bf{y}}_{i-1}^{I}$ 
and the aligned encoder hidden state ${\bf{e}}_{i}$:

\begin{equation}
{\bf{h}}_ { i }^{I} = f \left( {\bf{h}} _ { i - 1 }^{I} , {\bf{y}} _ { i - 1 }^{I} , {\bf{e}} _ { i }  \right).
\end{equation}
Then the decoder state $\mathbf{h}_{i}^
{I}$ is utilized for intent detection:
\begin{eqnarray}
{\bf{y}} _ { i } ^ { I } &=& \operatorname { softmax } \left( {\bf{W}} _ { h } ^ { I }{\bf{h}} _ { i }^{I} \right), \\
{o} _ { i } ^ { I } &=& \operatorname { argmax } ({\bf{y}} _ { i } ^ { I }),
\end{eqnarray}
where ${\bf{y}} _ { i } ^ { I }$ is the intent output distribution of the $i$th token in the utterance; ${{o}_{i}^{I}} $ represents the intent lable of $i$th token and ${\bf{W}}_{h}^{I}$ are trainable parameters of the model.

The final utterance result ${o}^{I}$ is generated by 
voting from all token intent results:
\begin{equation}
{{o}^{I}} =   \operatorname {argmax} {\sum _ { i = 1 } ^ {m } \sum _ { j = 1 } ^ { n_{I} } \alpha _ { j }  \mathds{1}[{{o}_{i}^{I}} = j] } ,
\end{equation}
where $m$ is the length of utterance and $n_{I}$ is the number of 
intent labels; $\alpha_{j}$ denotes a  0-1 vector  
$\alpha$  $\in$  $\mathbb{R}^{n_{I}}$ of which the $j$th unit is one and the 
others are zero;  $\operatorname {argmax}$ indicates the operation  returning 
the indices of the maximum values in $\alpha$.

By performing the token-level intent detection, there are mainly two advantages: 

1. Performing the token-level intent detection can provide features at each token to slot filling in our Stack-Propagation framework, which can ease the error propagation and retain more useful information for slot filling. Compared with sentence-level intent detection \cite{li2018self},
if the intent of the whole sentence is predicted wrongly, the wrong intent would possibly apply a negative impact on all slots. However, in token-level intent detection, if some tokens in the utterance predicted wrongly, other correct token-level intent information will still be useful for the corresponding slot filling.

2. Since each token can grasp the whole utterance contextual information by using the self-attentive encoder, 
we can consider predictions at each token in an utterance as individual prediction to the intent of this utterance.
And therefore, like ensemble neural networks, this approach will reduce the 
predicted variance and improve the performance of intent detection.
The experiments section empirically demonstrate the effectiveness of the token-level intent detection.

\subsection{Stack-propagation for Slot Filling}
In this paper, one of the advantages of our Stack-Propagation framework is directly
leveraging the explicit intent information to constrain the slots into a specific intent and alleviate the burden of slot filling decoder.
In our framework,
we compose the input units for slot filling decoder by 
concatenating the intent output distribution ${\bf{y}} _ { i } ^ { I } $ 
and the aligned encoder hidden state ${\bf{e}}_{i}$.

For the slot-filling decoder, we similarly use 
another unidirectional LSTM as the slot-filling decoder. 
At the decoding step $i$, the decoder state ${\bf{h}}_{i}^{S}$ 
can be formalized as:
\begin{equation}
{\bf{h}} _ { i }^{S} = f \left( {\bf{h}} _ { i - 1 }^{S} , {\bf{y}} _ { i - 1 }^{S} ,  {\bf{y}} _ { i } ^ { I } \oplus{{\bf{e}} _ { i }}\right),
\end{equation}
where ${\bf{h}}_{i-1}^{S}$ is the previous decoder state;  
${\bf{y}}_{i-1}^{S}$ is the previous emitted slot label distribution.

Similarily, the decoder state ${\bf{h}}_{i}^
{I}$ is utilized for slot filling:
\begin{eqnarray}
{\bf{y}} _ { i } ^ { S } &=& \operatorname { softmax } \left( {\bf{W}} _ { h } ^ { S } {\bf{h}}_ { i }^{S}\right),\\
{o} _ { i } ^ { S } &=& \operatorname { argmax } ({\bf{y}} _ { i } ^ { S }),
\end{eqnarray}
where ${o} _ { i } ^ { S }$ is the slot label of the $i$th word in the utterance.
\subsection{Joint Training}
Another major difference between existing joint work \cite{zhang2016joint, goo2018slot} and our framework is the training method for intent detection,
where we convert
the sentence-level classification task into
token-level prediction to directly leverage
token-level intent information for slot filling.
And the intent detection objection is formulated as:

\begin{equation}
\mathcal { L } _ { 1 } \triangleq - \sum _ { j = 1 } ^ { m } \sum _ { i = 1 } ^ { n_{I} }  \hat { {\bf{y}} } _ { j } ^ { i,I } \log \left( {\bf{y}} _ { j } ^ {  i,I } \right).
\end{equation}
Similarly, the slot filling task objection is defined as: 
\begin{equation}
\mathcal { L } _ { 2 } \triangleq - \sum _ { j = 1 } ^ { m } \sum _ { i = 1 } ^ { n_{S} } { \hat { {\bf{y}} } _ { j } ^ { i, S } \log \left( {\bf{y}} _ { j } ^ { i,S } \right)},
\end{equation}
where ${\hat { {\bf{y}} } _ { j } ^ { i,I } }$ and 
$ {\hat { {\bf{y}}} _ { j } ^ { i,S } }$ are the gold intent label and 
gold slot label separately; $n_{S}$ is the number of slot labels.

To obtain both slot filling and intent detection jointly, 
the final joint objective is formulated as
\begin{equation}
\mathcal { L } _ { \theta } = \mathcal{ L }_{1} + \mathcal{ L }_{2}.
\end{equation}

Through the joint loss function, the shared 
representations learned by the shared self-attentive encoder can 
consider two tasks jointly and further ease the error propagation compared with pipeline models \cite{zhang2016joint}. 

\section{Experiments}
\subsection{Experimental Settings}
To evaluate the efficiency of our proposed model, we 
conduct experiments on two benchmark datasets. One
is the publicly ATIS dataset \cite{hemphill1990atis} 
containing audio recordings of flight reservations, and the other is the custom-intent-engines
collected by Snips (SNIPS dataset) \cite{coucke2018snips}.
\footnote{\url{https://github.com/snipsco/nlu-benchmark/tree/master/2017-06-custom-intent-engines}}
Both datasets used in 
our paper follows the same format and partition as in \newcite{goo2018slot}. 
The dimensionalities of the word embedding is 256 for ATIS dataset and 512 for SNIPS dataset. The self-attentive encoder
hidden units are set as 256. L2 regularization is used 
on our model is $1\times 10^{-6}$ and dropout ratio
is adopted is 0.4 for reducing overfit.
We use Adam \cite{kingma-ba:2014:ICLR} to 
optimize the parameters in our model and adopted 
the suggested hyper-parameters for optimization. 
For all the experiments, we select the model which works the best on the dev set, and then evaluate it on the test set.
\subsection{Baselines}
We compare our model with the existing  
baselines including:
\begin{itemize}
	
	\item \textbf{Joint Seq}. \newcite{hakkani2016multi} 
	proposed a multi-task modeling approach for jointly modeling domain detection, intent detection, and slot filling in a single recurrent neural network (RNN) architecture. 
	
	\item \textbf{Attention BiRNN}. \newcite{liu2016attention} leveraged the 
	attention mechanism to allow the network to learn
	the relationship between slot and intent.
	
	\item \textbf{Slot-Gated Atten}. \newcite{goo2018slot} proposed the slot-gated joint model to explore the correlation of slot filling and intent detection better. 
	
	\item \textbf{Self-Attentive Model}. \newcite{li2018self} proposed 
	a novel self-attentive model with the intent augmented gate mechanism
	to utilize the semantic correlation between slot and 
	intent.
	
	\item \textbf{Bi-Model}. \newcite{wang2018bi} proposed the 
	Bi-model to consider  the intent and slot 
	filling cross-impact to each other.
	
	\item \textbf{CAPSULE-NLU.} \newcite{zhang-etal-2019-joint} proposed a capsule-based neural network model with a dynamic routing-by-agreement schema to accomplish slot filling and intent detection.
	
	\item \textbf{SF-ID Network.}  \cite{e-etal-2019-novel} introduced an SF-ID network to establish direct connections for the slot filling and intent detection to help them promote each other mutually.
	
\end{itemize}
For the \textit{Joint Seq}, \textit{Attention BiRNN}, \textit{Slot-gated Atten}, \textit{CAPSULE-NLU} and \textit{SF-ID Network}, we adopt the reported results from \newcite{goo2018slot, zhang-etal-2019-joint,e-etal-2019-novel}. For the \textit{Self-Attentive Model}, \textit{Bi-Model}, we re-implemented the models and obtained the results on the same datasets.\footnote{All experiments are conducted on the publicly datasets provided by \newcite{goo2018slot}, \textit{Self-Attentive Model} and \textit{Bi-Model} don't have the reported result on the same datasets or they did different preprocessing. For directly comparison, we re-implemented the models and obtained the results on the ATIS and SNIPS datasets preprocessed by \citet{goo2018slot}. Because all baselines and our model don't apply CRF layer, we just report the best performance of \textit{SF-ID Network} without CRF. It's noticing that our model does outperform \textit{SF-ID Network} with CRF layer.}

\begin{table*}[th!]
	\centering
	\resizebox{1.0\textwidth}{!}{
		\begin{tabular}{l|ccc|ccc}
			\hline
			\multirow{2}{*}{\textbf{Model}} & \multicolumn{3}{c|}{\textbf{SNIPS}} & \multicolumn{3}{c}{\textbf{ATIS}} \\ \cline{2-7} 
			& Slot (F1) &Intent (Acc) & Overall (Acc) 
			& Slot (F1) &Intent (Acc) & Overall (Acc)           \\ 
			\hline
			Joint Seq \cite{hakkani2016multi}  
			&87.3	&96.9	&73.2&94.3   &92.6	&80.7 	\\
			Attention BiRNN \cite{liu2016attention} &87.8	&96.7	&74.1&94.2   &91.1	&78.9		\\
			
			Slot-Gated Full Atten \cite{goo2018slot}	&88.8	&97.0	&75.5&94.8   &93.6	&82.2		\\
			Slot-Gated Intent Atten \cite{goo2018slot}	&88.3   &96.8	&74.6&95.2	&94.1	&82.6	 	\\
			Self-Attentive Model \cite{li2018self}
			& 90.0 &  97.5  & 81.0	& 95.1		&96.8  	& 	82.2   		\\
			Bi-Model \cite{wang2018bi}	  & 93.5 &  97.2  & 83.8& 95.5		&96.4  	& 	85.7		\\
			CAPSULE-NLU \cite{zhang-etal-2019-joint}	  & 91.8 &  97.3  & 80.9& 95.2		&95.0  	& 	83.4		\\
			 SF-ID Network \cite{e-etal-2019-novel}	  & 90.5 &  97.0  & 78.4& 95.6		&96.6  	& 	86.0		\\
			\hdashline
			
			Our model   &\textbf{       94.2*} &\textbf{ 98.0*}	&\textbf{ 86.9*} &\textbf{ 95.9*}	& \textbf{ 96.9*}	&\textbf{ 86.5*} 	\\ 
			Oracle (Intent)  &{96.1} &{-}	&{-}&{96.0}	& {-}	&{-} \\ 
			\hline
		\end{tabular}
	}
	\vspace{0.05in}
	\caption{Slot filling and intent detection results on two datasets.   The numbers with * indicate that the improvement of our model over all baselines
		is statistically significant with $p<0.05$ under t-test.}
	\label{table:overall_result}
	
\end{table*}
\subsection{Overall Results}\label{sec:result:all}
Following \newcite{goo2018slot}, we evaluate the 
SLU performance of slot filling using 
F1 score and the performance of intent prediction using accuracy, 
and sentence-level semantic frame parsing 
using overall accuracy.
Table~\ref{table:overall_result} shows the
experiment results of the proposed models
on SNIPS and ATIS datasets.

From the table, we can see that our model
significantly outperforms all the baselines by a 
large margin and achieves the state-of-the-art 
performance.  In the SNIPS dataset, compared 
with the best prior joint work \textit{Bi-Model},
we achieve 0.7\% improvement on Slot (F1) score, 
0.8\% improvement on Intent (Acc) and 3.1\% 
improvement on Overall (Acc). In the ATIS dataset, we 
achieve 0.4\% improvement on Slot (F1) score, 
0.5\% improvement on Intent (Acc) and 0.8\%  
improvement on Overall (Acc).
This indicates the effectiveness of our 
Stack-Propagation framework. Especially, our framework gains the 
largest improvements on sentence-level semantic frame accuracy, we attribute this to the fact 
that our framework directly take the explicit intent information into 
consideration can better help grasp the relationship between the intent and slots and improve the SLU performance.

To see the role of intent information for SLU tasks intuitively, we also present the result when using the gold intent 
information.\footnote{During the inference time, we concatenate the gold intent 
	label distribution (one-hot vector) and the aligned encoder hidden state  $e_{i}$ as the 
	composed input for slot filling decoder. To keep the train and test procedure as the same, we replace our intent distribution as one-hot intent information for slot filling when train our model in our oracle experiments setting.} The result is 
shown in the \textit{oracle} row of Table~\ref{table:overall_result}. 
From the result, we can see that further leveraging better intent
information will lead to better slot filling performance. The result also verifies our assumptions that intent information can be used for guiding the slots prediction.

\subsection{Analysis}

In Section 4.3, significant improvements among all three metrics have been witnessed on both two publicly datasets. However, we would like to know the reason for the improvement. In this section, 
we first explore the effect of Stack-Propagation framework.
Next, we study the effect of our proposed token-level intent detection mechanism.
Finally, we study the effect of self-attention mechanism in our framework. 
\begin{table*}[th!]
	\centering
	\resizebox{1.0\textwidth}{!}{
		\begin{tabular}{l|ccc|ccc}
			\hline
			\multirow{2}{*}{\textbf{Model}} & \multicolumn{3}{c|}{\textbf{SNIPS}} & \multicolumn{3}{c}{\textbf{ATIS}} \\ \cline{2-7} 
			& Slot (F1) &Intent (Acc) & Overall (Acc) 
			& Slot (F1) &Intent (Acc) & Overall (Acc)           \\ 
			\hline
			gate-mechanism  
			&92.2	&97.6	&82.4& 95.3&  96.2 & 83.4  	\\
			pipelined model 
			&90.8	&97.6	&81.8 &95.1 &96.1	&82.3	\\

			sentence intent augmented	&93.7   &97.5	&86.1&95.5	&96.7	&85.8	 
			\\ 
			lstm+last-hidden & - &  97.1  &-  & -		&95.2  	& 	-		\\
			lstm+token-level 	   &- &  97.5  & - & -		&96.0	& 	- 		\\
				without self-attention	&94.1	&97.8	&86.6&95.6   &96.6	&86.2		\\
			\hdashline
			
			Our model    &\textbf{94.2} &\textbf{98.0}	&\textbf{86.9}&\textbf{95.9}	& \textbf{96.9}	&\textbf{86.5} 	\\ 
			
			\hline
		\end{tabular}
	}
	\vspace{0.05in}
	\caption{The SLU performance on baseline models compared with our Stack-Propagation model on two datasets.}
	\label{tab:ppn}
\end{table*}
\subsubsection{Effect of Stack-Propagation Framework}
To verify the effectiveness of the Stack-Propagation framework.
we conduct experiments with the following ablations:

1) We conduct experiments to incorporate intent information by using gate-mechanism which is similar to \newcite{goo2018slot}, providing the intent information by interacting with the slot filling decoder by gate function.\footnote{For directly comparison, we still perform the token-level intent detection.} We refer it as \textit{gate-mechanism}.

2) We conduct experiments on the pipelined model where the intent detection and slot filling has their own self-attentive encoder separately. The other model components keep the same as our framework.
We name it as \textit{pipelined model}.

Table \ref{tab:ppn} 
gives the result of the comparison experiment.
From the result of \textit{gate-mechanism} row, we can observe that 
without the Stack-Propagation learning and simply using the gate-mechanism to incorporate the intent information,
the slot filling (F1) performance drops 
significantly, which demonstrates that directly 
leverage the intent information with Stack-Propagation can improve the slot filling performance effectively than using the gate mechanism.
Besides, we can see that the intent detection (Acc) and overall accuracy (Acc) decrease a lot.
We attribute it to the fact 
that the bad slot filling performance harms 
the intent detection and the whole sentence 
semantic performance due to the joint learning scheme.

Besides, from the \textit{pipeline model} row of Table~\ref{tab:ppn},
we can see that without shared encoder, the performance on all metrics 
declines significantly. 
This shows that Stack-Propagation model can learn the correlation knowledge which may promote each other and ease the error propagation effectively.

\subsubsection{Effect of Token-Level Intent Detection Mechanism}
In this section, we study the effect of the proposed token-level intent 
detection with the following ablations:

1) We conduct the sentence-level intent detection 
in intent detection separately, which utilizes the last hidden vector of BiLSTM 
encoder for intent detection. We refer it to \textit{lstm + last-hidden}.
For comparison, our token-level intent detection without joint learning with slot filling is named as \textit{lstm+token-level} in Table~\ref{tab:ppn}.

2) We conduct a joint learning framework that slot filling uses the
utterance-level intent information rather than token-level intent information for each token, similar to intent-gated 
mechanism \cite{li2018self}, which is named as \textit{sentence intent augmented.}

We show these two comparison experiments results in the first block of 
table~\ref{tab:ppn}. From the result, we can see that the 
token-level intent detection obtains better performance than the 
utterance-level intent detection. We believe the reason is that intent prediction on each token has similar advantage to ensemble neural networks, which can reduce the predicted variance to improve the intent performance. 
As a result, our framework can provide more useful intent information for slot filling by introducing token-level intent detection.

In addition, we can observe that if we only 
provide the sentence-level intent information for slot filling
decoder, we obtain the worse results, which demonstrates the 
significance and effectiveness of incorporating token-level 
intent information. The main reason for this can be that
incorporating the token-level intent information can retain 
useful features for each token and ease the error propagation.

\subsubsection{Effect of Self-attention Mechanism}
We further investigate the benefits of self-attention mechanism in our framework. We 
conduct the comparison experiments with the same framework except the self-attention encoder is
replaced with BiLSTM. 

Results are shown in the \textit{without self-attention} row of Table~\ref{tab:ppn}. 
We can observe the self-attention 
mechanism can further improve the SLU performance. We attribute this to the fact that self-attention 
mechanism can capture the contextual information for each token. 
Without the self-attention mechanism, it will harm the intent detection and have bad
influence on slot filling task by joint learning.

It is noticeable that even without the self-attention mechanism, our framework still performs the state-of-the-art Bi-model model \cite{li2018self}, which again demonstrates the effectiveness and robustness of 
our other framework components.
\begin{table*}[th!]
	\centering
	\resizebox{1.0\textwidth}{!}{
		\begin{tabular}{l|ccc|ccc}
			\hline
			\multirow{2}{*}{\textbf{Model}} & \multicolumn{3}{c|}{\textbf{SNIPS}} & \multicolumn{3}{c}{\textbf{ATIS}} \\ \cline{2-7} 
			& Slot (F1) &Intent (Acc) & Overall (Acc) 
			& Slot (F1) &Intent (Acc) & Overall (Acc)           \\ 
			\hline

			Our model    &{94.2} &{98.0}	&{86.9}&{95.9}	& {96.9}	&{86.5} 	\\ 
			\hdashline
			Intent detection (BERT) &{-} &{97.8}	&{-}&{-}	& {96.5}	&{-}	\\ 
			Slot filling (BERT) &{95.8} &{-}	&{-}&{95.6}	& {-}	&{-}	\\ 
			\hdashline
			BERT SLU \cite{chen2019bert} &{97.0} &{98.6}	&{92.8}&{96.1}	& {97.5}	&{88.2}	\\ 
			Our model + BERT &{97.0} &{99.0}	&{92.9}&{96.1}	& {97.5}	&{88.6}	\\ 
			\hline
		\end{tabular}
	}
	\vspace{0.05in}
	\caption{The SLU performance on BERT-based model on two datasets.}
	\label{tab:bert}
\end{table*}
 
\subsection{Effect of BERT}
Finally, we also conduct experiments to use pre-trained model, BERT \cite{devlin2018bert}, to boost SLU performance. 
In this section, we replace the self-attentive encoder by \textit{BERT base} model
with the 
fine-tuning approach and keep other components as same with our framework.

Table~\ref{tab:bert} gives the results of BERT 
model on ATIS and SNIPS datasets. From the table, the
BERT model performs remarkably well on both two
datasets and achieves a new state-of-the-art performance, which
indicates the effectiveness of a strong pre-trained model in 
SLU tasks. We attribute this to the fact that 
pre-trained models can provide rich semantic features,
which can help to improve the performance on SLU tasks.
In addition, our \textit{model + BERT} outperforms the \textit{BERT SLU} \cite{chen2019bert} which apply BERT for joint the two tasks and there is no explicit interaction between intent detection and slot filling in two datasets in overall acc metric. It demonstrates that our framework is effective with BERT. 

Especially, we also conduct experiments of intent detection task and slot filling separately based on BERT model. For intent detection,
we put the special \texttt{[CLS]} word embedding into a classification layer to classify the intent. For slot filling, we
feed the final hidden representation $\mathbf{h_{\text{BERT}}}_{i}$ $\in$ $\mathbb{R}^{d}$ for each token $i$ \footnote{We only consider the first subword label if a word is broken into multiple subwords} into a classification layer 
over the slot tag set. 
The results are also shown in the Table~\ref{tab:bert}. From the result, we can see that the slot filling 
(F1) and intent detection accuracy (Acc) is lower than our joint model based on BERT, which again demonstrates the effectiveness of exploiting the relationship between these two tasks. 

\section{Related Work}
Slot filling can be treated as a sequence labeling task, and the popular approaches are conditional random fields (CRF) \cite{raymond2007generative} and recurrent neural networks (RNN) \cite{xu2013convolutional,yao2014spoken}. 
The intent detection is formulated as an utterance classification problem,
and different classification methods, such as support vector machine (SVM) and RNN \cite{haffner2003optimizing,sarikaya2011deep},  has proposed to solve it.

Recently, there are some joint models to overcome the error propagation caused by the pipelined approaches.
\newcite{zhang2016joint} 
first proposed the joint work using RNNs for learning the correlation between intent and slots. \newcite{hakkani2016multi} 
proposed a single recurrent neural network for modeling slot filling
and intent detection jointly. \newcite{liu2016attention} 
proposed an attention-based neural network
for modeling the two tasks jointly. All these models outperform the pipeline models via mutual enhancement between two tasks. 
However, these joint models
did not model the intent information for slots explicitly and just considered their correlation between the two tasks by sharing parameters.

Recently, some joint models have explored incorporating the intent information for slot filling.
\newcite{goo2018slot} utilize a slot-gated mechanism as a special gate function to model the relationship between
the intent detection and slot filling.
\newcite{li2018self} proposed the intent augmented gate mechanism to utilize the semantic correlation between slots and intent. 
Our framework is significantly different from their models including: (1)  both of their approaches utilize the gate mechanism to model the relationship
between intent and slots. While in our model, to directly leverage the intent information in the joint model,
we feed the predicted intent information directly into slot filling with Stack-Propagation framework.
(2) They apply the sentence-level intent information for each word while we adopt the token-level intent information for slot filling and further ease the error propagation.
\newcite{wang2018bi}
propose the Bi-model to consider the cross-impact between the intent and slots and achieve the state-of-the-art result. 
\newcite{zhang-etal-2019-joint} propose a hierarchical capsule neural network to model the the hierarchical relationship among word, slot, and intent in an utterance.
\newcite{e-etal-2019-novel} introduce an SF-ID network to establish the interrelated mechanism for slot filling and intent detection tasks.
Compared with their works,  our model can directly incorporate the intent 
information for slot filling explicitly with Stack-Propagation which makes the interaction procedure more interpretable, while their model just interacts with hidden state implicitly between two tasks.

\section{Conclusion}
In this paper, we propose a joint model for spoken language understanding with  
Stack-Propagation to
better incorporate the intent information for slot filling.
In addition, we perform the token-level intent detection to improve the intent detection performance and further ease the error propagation.
Experiments on two datasets show the effectiveness 
of the proposed models and achieve the state-of-the-art 
performance. Besides, we explore and analyze the effect
of incorporating strong pre-trained BERT model in SLU tasks. With BERT, the result reaches a new state-of-the-art level.

\section*{Acknowledgments}
We thank the anonymous reviewers for their helpful comments and suggestions.
This work was supported by the National Natural Science Foundation of China (NSFC) via grant 61976072, 61632011 and 61772153.

\bibliography{emnlp-ijcnlp-2019}
\bibliographystyle{acl_natbib}

\appendix

\end{document}